\documentclass[sigconf]{acmart}
\makeatletter
\def\@ACM@checkaffil{
    \if@ACM@instpresent\else
    \ClassWarningNoLine{\@classname}{No institution present for an affiliation}%
    \fi
    \if@ACM@citypresent\else
    \ClassWarningNoLine{\@classname}{No city present for an affiliation}%
    \fi
    \if@ACM@countrypresent\else
        \ClassWarningNoLine{\@classname}{No country present for an affiliation}%
    \fi
}
\makeatother
\usepackage{hyperref}
\usepackage{algorithm}
\usepackage{subcaption}
\usepackage{algpseudocode}
\usepackage{xspace}
\usepackage{array}
\usepackage{ragged2e} 
\usepackage[normalem]{ulem}  
\usepackage{titlesec}
\titlespacing*{\section}{0pt}{0.1\baselineskip}{0.1\baselineskip} 
\AtBeginDocument{%
  }
\setcopyright{acmcopyright}
\copyrightyear{2023}
\acmYear{2023}
\setcopyright{rightsretained}
\acmConference[CIKM '23]{Proceedings of the 32nd ACM International Conference on Information and Knowledge Management}{October 21--25, 2023}{Birmingham, United Kingdom}
\acmBooktitle{Proceedings of the 32nd ACM International Conference on Information and Knowledge Management (CIKM '23), October 21--25, 2023, Birmingham, United Kingdom}\acmDOI{10.1145/3583780.3615262}
\acmISBN{979-8-4007-0124-5/23/10}
\definecolor{orange}{RGB}{255,119,0}
\definecolor{red}{RGB}{220,0,0}
\definecolor{agreen}{RGB}{74, 198, 148}
\definecolor{purple}{RGB}{158, 62, 177}
\definecolor{darkpurple}{RGB}{170, 70, 210}
\definecolor{aqua}{RGB}{87, 180, 181}
\definecolor{lightblue}{RGB}{72, 123, 232}
\definecolor{hotpink}{RGB}{255, 83, 115}
\definecolor{teal}{RGB}{90, 200, 250}
\definecolor{linkColor}{RGB}{6,125,233}
\definecolor{tomato}{rgb}{1,0.2,0}
\definecolor{grey}{rgb}{0.4,0.4,0.4}

\DeclareMathOperator*{\argmin}{arg\,min}

\newcolumntype{C}[1]{>{\centering\arraybackslash}p{#1}}
\newcolumntype{R}[1]{>{\RaggedLeft\arraybackslash}p{#1}}
\newcolumntype{L}[1]{>{\RaggedRight\arraybackslash}p{#1}}
\begin{document}

\title{Patient Clustering via Integrated Profiling of Clinical and Digital Data
}
\author{Dongjin Choi}
\orcid{0000-0002-7311-0644}
\affiliation{%
  \institution{Georgia Institute of Technology, USA}
}
\email{jin.choi@gatech.edu}

\author{Andy Xiang}
\orcid{0009-0008-4918-0221}
\affiliation{%
  \institution{Kaiser Permanente, USA}
}
\email{Andy.X.Xiang@kp.org}

\author{Ozgur Ozturk}
\orcid{0009-0000-0109-2419}
\affiliation{%
  \institution{Kaiser Permanente, USA}
}
\email{Ozgur.X.Ozturk@kp.org}

\author{Deep Shrestha}
\orcid{0009-0001-8775-7596}
\affiliation{%
  \institution{Kaiser Permanente, USA}
}
\email{deep.x.shrestha@kp.org}

\author{Barry Drake}
\orcid{0000-0003-4087-1524}
\affiliation{%
  \institution{Georgia Tech Research Institute, USA}
}
\email{drakeleeb@gmail.com}

\author{Hamid Haidarian}
\orcid{0009-0005-6299-7875}
\affiliation{%
  \institution{Kaiser Permanente, USA}
}
\email{Hamid.Haidarian@kp.org}

\author{Faizan Javed}
\orcid{0009-0003-7075-5097}
\affiliation{%
  \institution{Kaiser Permanente, USA}
}
\email{Faizan.X.Javed@kp.org}

\author{Haesun Park}
\orcid{0000-0001-6259-7170}
\affiliation{%
  \institution{Georgia Institute of Technology, USA}
}
\email{hpark@cc.gatech.edu}

\renewcommand{\shortauthors}{Dongjin Choi et al.}

\begin{abstract}
We introduce a novel profile-based patient clustering model designed for clinical data in healthcare. By utilizing a method grounded on constrained low-rank approximation, our model takes advantage of patients' clinical data and digital interaction data, including browsing and search, to construct patient profiles. As a result of the method, nonnegative embedding vectors are generated, serving as a low-dimensional representation of the patients.
Our model was assessed using real-world patient data from a healthcare web portal, with a comprehensive evaluation approach which considered clustering and recommendation capabilities.
In comparison to other baselines, our approach demonstrated superior performance in terms of clustering coherence and recommendation accuracy.
\end{abstract}




\maketitle

\begin{figure}[tb]
  \centering
  \includegraphics[width=0.45\textwidth]{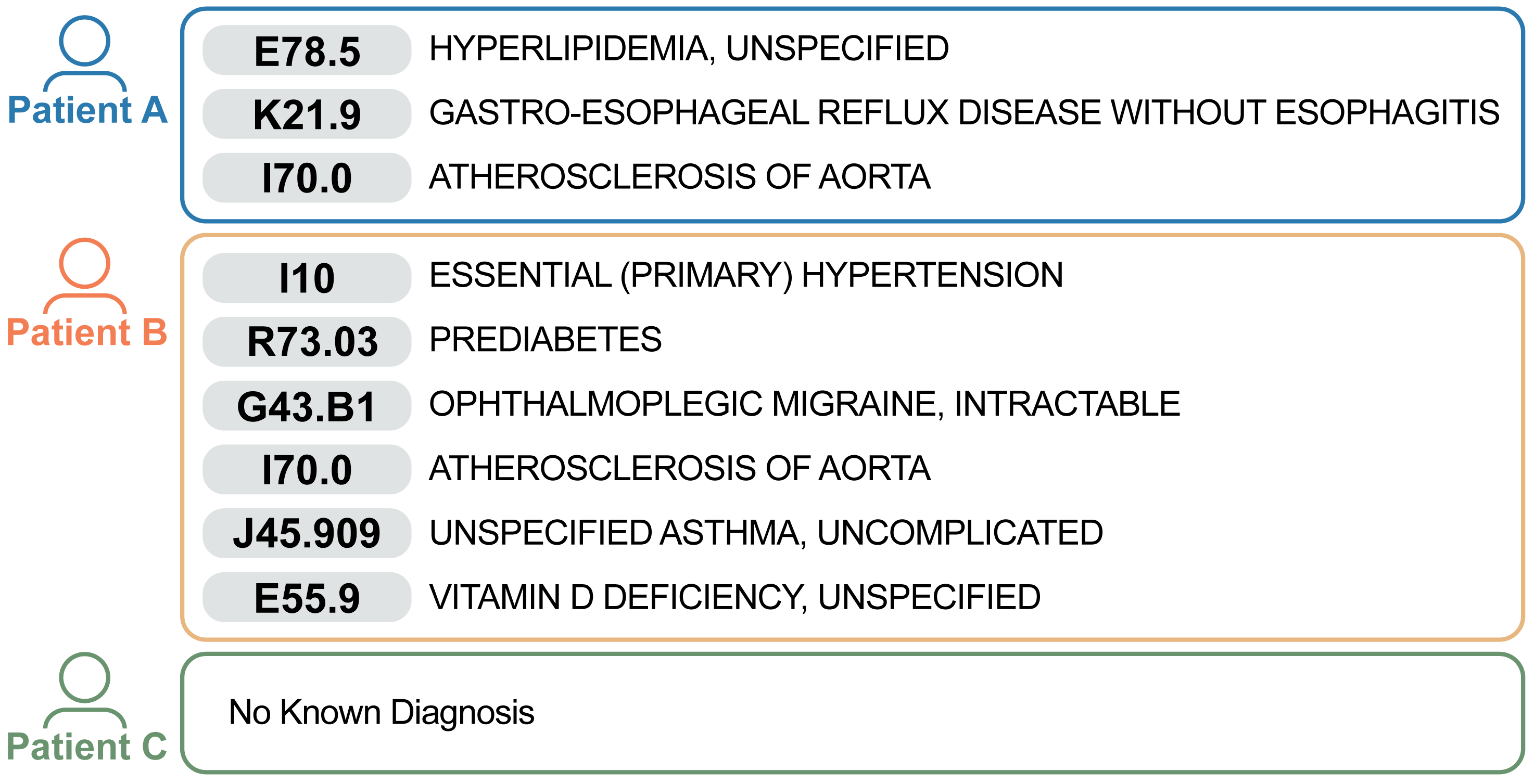}
  \vspace{-3mm}
  \caption{Example clinical data of hypothetical patients. Each box lists the diagnostic codes and corresponding conditions for the respective patient. Patient C is currently undiagnosed.}
  \vspace{-3mm}
  \label{fig:dx_example}
\end{figure}
\begin{figure*}[tb]
  \centering
  \includegraphics[width=0.92\textwidth]{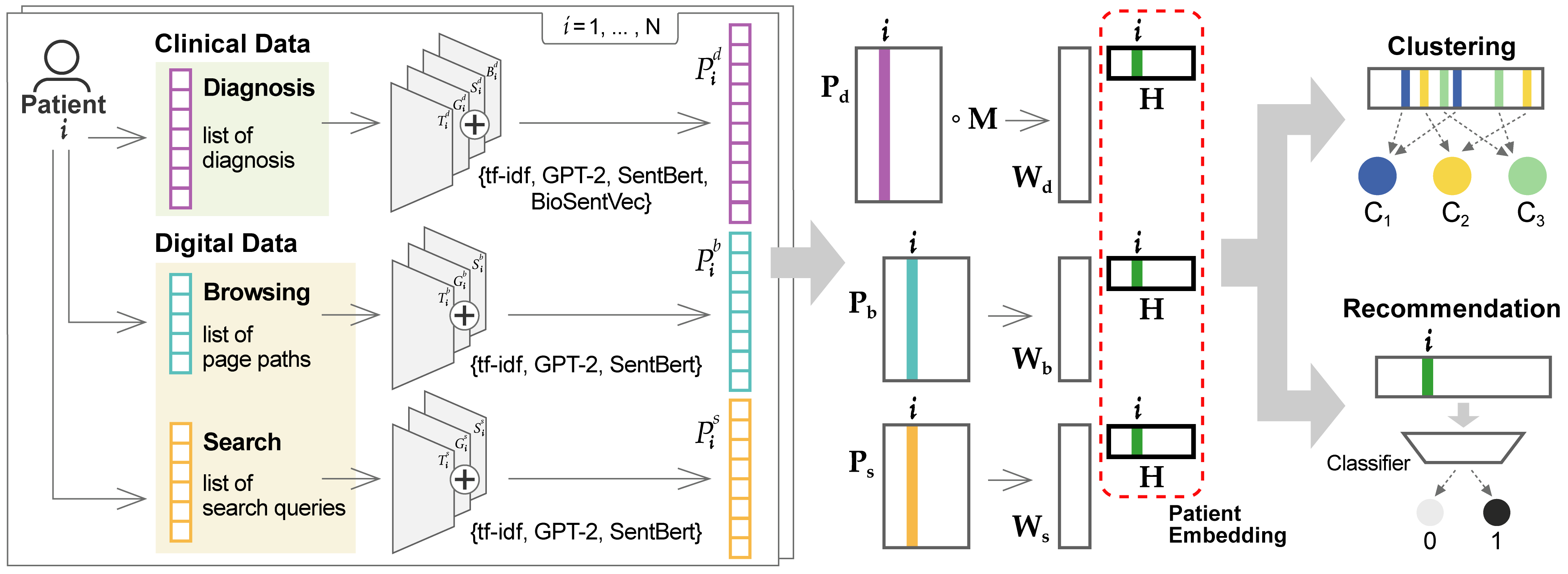}
  \vspace{-3mm}
  \caption{Illustration of the proposed patient profiling and clustering framework.}
  \vspace{-3mm}
  \label{fig:overview}
\end{figure*}
\section{Introduction}
Clinical data, particularly ICD-10-CM codes~\cite{world1992icd}, is pivotal for monitoring patient health as depicted in \autoref{fig:dx_example}, a hypothetical example clinical data.
A significant challenge is the effective clustering of patients based on these complex and multi-dimensional data.
In addition, with the advancement of healthcare web portals, clinical data is becoming increasingly connected with digital data. Beyond traditional clinical data records, these web portals record user activities that supplement the traditional clinical data records, such as browsing web pages, searching for general information, and acquiring doctor and clinical information~\cite{tanbeer2021myhealthportal,coughlin2017patient,chou2002healthcare,moody2005health}. The integration of such digital data into healthcare analysis has the potential to enhance the accuracy of patient clustering and recommendations.
Our work uniquely integrates clinical data with digital data, surpassing traditional profile-building. Instead of focusing solely on clinical clustering or digital data, we unify both, pioneering new representation techniques.

Profile-based models have long been used significantly in clustering and recommendation, with variants developed for various domains and data~\cite{DBLP:conf/ictai/PretschnerG99,DBLP:conf/cikm/ShenTZ05,DBLP:conf/www/SugiyamaHY04,DBLP:conf/kdd/TanSZ06}.
They derive user profiles from system interaction histories, linking this information to preference scores for recommendations items.
Profile-based methods largely overlap with representation learning or embedding models that are gaining widespread use in diverse fields.
Although embedding models often focus on using the learned vector in downstream machine learning models regardless of numerical values~\cite{DBLP:conf/nips/MikolovSCCD13}, profile-based models assume that each dimension of the learned profile vectors holds latent significance, utilizing this for further recommendations.

Our research integrates clinical data and digital data to better understand their health status. First, we utilize the textual information associated with the records. For categorical data like ICD-10-CM codes, we utilize expert-annotated description text. For browsing and search records, we extract textual content, including queries and page paths. Second, we employ pre-trained embedding methods for textual representation, using domain-specific BioSentVec~\cite{chen2019biosentvec}, and general models like GPT-2~\cite{radford2019language} and SentenceBERT~\cite{reimers2019sentence} for contextual semantics.

For patient clustering and embedding, we introduce an algorithm that uses low-rank approximation with nonnegativity constraints. Drawing from prior nonnegative matrix factorization research~\cite{kim2008nonnegative,kim2011fast}, nonnegativity enhances result interpretability aligning with soft clustering~\cite{kuang2015nonnegative}. 
It also learns a lower-dimensional representation of patients, 
providing a probabilistic interpretation of the results~\cite{ekstrand2011collaborative}. 

We conducted extensive experiments to evaluate our model using real-world data from the web portal of our collaborator, Kaiser Permanente. 
Our results demonstrate that our method outperforms other approaches in terms of clustering coherence and recommendation accuracy.
\section{Related Work}
Our study integrates research areas such as clinical data clustering, as well as data embedding methods and advanced mathematical approaches like constrained low rank approximation and nonnegative matrix factorization.
Clinical data clustering involves grouping diagnosis or patient data points based on their similarity. This is also highly related to embedding learning and is becoming important as it aids in accurate diagnosis prediction and recommendation~\cite{henriksson2015modeling,choi2016retain,tan2022metacare++}.
Moreover, there has been work on leveraging multiple dimensions of patient data, such as educational level, health literacy, and emotional status, to enhance recommendation techniques~\cite{stratigi2020multidimensional}.
Our research employs advanced mathematical techniques, including constrained low rank approximation (CLRA), and nonnegative matrix factorization (NMF). These methodologies have proven beneficial for devising efficient clustering techniques, allowing learning of a lower-dimensional representation at the same time~\cite{du2019hybrid,SymNMF_Jogo,HSymNMF2,whang2020mega,kim2008sparse,kim2015simultaneous}. Our approach combines mathematical techniques with data integration for innovative patient clustering. We bridge individual data types for a unified representation, distinguishing our work from traditional methods.
\section{Proposed Method}
\subsection{Problem Definition}
We aim to cluster patients utilizing their clinical and digital data. We define the diagnostic data $\mathcal{D}$ for patients. For each patient $p_i$, the set of current diagnoses is denoted by $D_i = \{d_{i,j}\}_{j=1}^{N_i^d}$, with $N_i^d$ as the diagnosis count for patient $i$. A diagnosis $d_{i,j}$ is expressed as an ICD-10-CM code, as shown in
\autoref{fig:dx_example} with examples of hypothetical patients. Patients A and B have diagnoses, while Patient C is currently undiagnosed. Each code is associated with a text description annotated by the Centers for Disease Control and Prevention (CDC)\footnote{\url{https://www.cdc.gov/nchs/icd/}}, e.g., R73.03 is for prediabetes. 
Diagnoses are sparse; in our real-world data of 18,000 possible codes, patients typically had only 10 diagnoses on average, with 5\% having no recorded diagnoses.
We propose a patient profiling technique to address the data sparsity and unobserved sets, using text embedding and processing methods.

\vspace{-1mm}
\subsection{Constructing Patient Profiles}
Our method's architecture is detailed in \autoref{fig:overview}.
We construct a diagnosis profile $P_i^{d}$ for patient $i$ using various embedding techniques and text processing methods.
First, a partial profile $T_i^{d}$ is constructed using TF-IDF (term frequency-inverse document frequency) scheme~\cite{DBLP:reference/ml/2010}. The $j^{th}$ element of $T_i^{d}$, $(T_i^{d})_j$, is computed as:
\vspace{-1mm}
\begin{equation}
    (T_i^{d})_j = \text{tf-idf}_{d}(t_j, i),
\end{equation}
where $\text{tf-idf}_{d}(t_j, i)$ is the tfidf score for term $t_j$ in patient $i$'s diagnosis description. 
Next, we employ BioSentVec~\cite{chen2019biosentvec}, a clinical domain-centric matrix-factorization-based embedding, which outperforms general embeddings like Word2Vec~\cite{mikolov2013distributed} and Doc2Vec~\cite{le2014distributed} in this context. Given that highly domain-specific terms exist in the diagnosis descriptions, BioSentVec is crucial in understanding the clinical context.
Patient diagnosis records are one-hot encoded, e.g. $[1, 0, 0, 1]$.
For instance, if a patient's diagnosis set is [F33.1, G89.21], the slots for F33.1 and G89.21 are set to 1s, while others are 0s. We then compute $B_i^{d}$ as follows:
\begin{equation}
B_i^{d} = \frac{\mathbf{E}_{B} O_i^{d}}{\sum(O_i^{d})},
\end{equation}
where $O_i^{d}\in \{0, 1\}^{n_d\times 1}$ is a binary vector with a length of number of all diagnosis, $n_d$. $\mathbf{E}_{B} \in \mathbb{R}^{d_B \times n_d}$ stacks the BioSentVec column embeddings for all diagnosis descriptions. Therefore, $B_i^{d}$ is the average of the BioSentVec embeddings for diagnoses that patient $i$ holds.

To further understand the context, we utilize GPT-2~\cite{radford2019language} and SentenceBERT~\cite{reimers2019sentence}. Similar to $\mathbf{E}_B$ for BioSentVec, we compute stacked matrices of embeddings from GPT-2 and SentenceBERT, denoted as $\mathbf{E}_{G}$ and $\mathbf{E}_{S}$, respectively. This enables the derivation of $G_i^{d}$ and $S_i^{d}$, sub-profiles for patient $i$, similarly. The intermediate steps for these computations are omitted for brevity.
Sub-profiles $T_i^{d}$, $G_i^{d}$, $B_i^{d}$, and $S_i^{d}$ are transformed using Min-Max scaling~\cite{DBLP:journals/pr/JainNR05} based on the values of the corresponding sub-profiles across all patients, and then concatenated to form the diagnosis profile for each patient:
\begin{equation}
P_i^d = \left[ T_i^{d} ; G_i^{d} ; S_i^{d} ; B_i^{d} \right],
\end{equation}
where `$;$' denotes the operation of vertical concatenation of vectors. 

Our method incorporates additional digital data to fully utilize available data sources and achieve higher accuracy. Motivated by the advancement of healthcare web portals, we propose algorithms that can effectively integrate digital data of user browsing and search activities to achieve more precise clustering of user profiles. To construct a user profile for browsing, denoted as $P_i^b$, and search, denoted as $P_i^s$, we employ a similar approach as calculating the diagnosis profile $P_i^d$ but using only TF-IDF, GPT-2, and SentenceBERT. 
Specifically, we represent the set of browsing activities for patient $i$, $p_i$, in the same way as the diagnosis, denoted as $B_i = \{p_{i,j}\}_{j=1}^{N_i^b}$. Each $p_{i,j}$ in the set represents a page path, the location of a web page visited by patient $i$ within the web portal. Similarly, we represent search activities as $S_i = \{q_{i,j}\}_{j=1}^{N_i^s}$, where $q_{i,j}$ denotes a query text the patient issued. We apply the same scaling and concatenation steps used in the process of constructing the diagnosis profile.

\subsection{Constrained Low Rank Approximation}
We assemble individual patient profiles into data matrices. 
We generate a diagnosis profile matrix $\mathbf{P}_d \in {\mathbb{R}}^{M_d \times N}$, a browsing profile matrix $\mathbf{P}_b \in {\mathbb{R}}^{M_b \times N}$, and a search profile matrix $\mathbf{P}_s \in {\mathbb{R}}^{M_s \times N}$, where
$N$ is the total number of patients. Specifically, the $i$-th column of $\mathbf{P}_d$, $\mathbf{P}_b$, and $\mathbf{P}_s$ corresponds to the diagnosis profile $P_i^d$, browsing profile $P_i^b$, and search profile $P_i^s$ of patient $i$, respectively.
We formulate an objective function for nonnegativity-constrained low rank approximation to minimize the discrepancy between original profile matrices ($\mathbf{P}_d, \mathbf{P}_b, \mathbf{P}_s$) and their respective low-rank approximations:
\begin{equation}
\label{eqn:merged}
\begin{split}
\min _{(\mathbf{W}_d, \mathbf{W}_b, \mathbf{W}_s, \mathbf{H})\geq 0} 
\left\|\mathbf{P}_d-\mathbf{W}_d \mathbf{H}\right\|_F^2 +
\alpha_b&\left\|\mathbf{P}_b -\mathbf{W}_b\mathbf{H}\right\|_F^2 \\
&+\alpha_s\left\|\mathbf{P}_s-\mathbf{W}_s \mathbf{H}\right\|_F^2,
\end{split}
\end{equation}
where $\alpha_b$ and $\alpha_s$ denote balancing factors for the low-rank approximation terms. Note that the factor $\textbf{H}$ ($\in \mathbb{R}_+^{k \times N}$) is common across all domains, and it provides a nonnegative embedding in $k$-dimensional space for patients.
The factors $\mathbf{W}_d$ ($\in \mathbb{R}_+^{M_d \times k}$), $\mathbf{W}_b$ ($\in \mathbb{R}_+^{M_b \times k}$), and $\mathbf{W}_s$ ($\in \mathbb{R}_+^{M_s \times k}$) represent the basis matrices in the reduced $k$-dimensional spaces.

We adopt a block coordinate descent (BCD) approach for optimization of ~\autoref{eqn:merged}.  
In each iteration of our proposed BCD method, we alternate updating one of the matrices $\mathbf{W}_d$, $\mathbf{W}_b$, $\mathbf{W}_s$, and $\mathbf{H}$ while fixing the other three matrices by solving the following subproblems until a stopping criteria is satisfied:
\begin{align}
\label{eqn:optimize_Ws}
\mathbf{W}_j &\leftarrow \argmin_{\mathbf{W}_j \geq 0} \left\|\mathbf{P}_j - \mathbf{W}_j\mathbf{H} \right\|_F , \quad \text{for}  \quad j \in \{d,b,s\}\\ 
\mathbf{H} &\leftarrow \argmin_{\mathbf{H} \geq 0} \left\|
\begin{bmatrix} \mathbf{P}_d \\
\sqrt{\alpha_b}\mathbf{P}_b \\
\sqrt{\alpha_s}\mathbf{P}_s \end{bmatrix} - \begin{bmatrix} \mathbf{W}_d \\
\sqrt{\alpha_b}\mathbf{W}_b \\
\sqrt{\alpha_s}\mathbf{W}_s \end{bmatrix}\mathbf{H}\right\|_F.
\label{eqn:optimize_H}
\end{align}
Each subproblem is a nonnegativity-constrained least squares (NLS) problem and we utilize the BPP (Block Principal Pivoting) method as it has been shown to produce the best performance in previous extensive studies \cite{kim2014algorithms}. Assuming that each subproblem has a unique solution, the limit point of the iteration is guaranteed to be a stationary point ~\cite{kim2014algorithms,bertsekas1997nonlinear}.

\subsection{Bypassing Unobserved Diagnosis}
Data and features are not always fully observed. We develop a method that properly handles unobserved or missing data and features.
For browsing and search data, we assume a closed-world assumption~\cite{DBLP:conf/adbt/Reiter77}, meaning that unobserved matrix entries indicate no existing relationship.
This is due to users having the freedom to browse and search, and they can also choose not to engage in such activities based on their intentions.
On the other hand, for diagnosis data, we utilize an open-world assumption~\cite{DBLP:journals/pieee/Nickel0TG16}, i.e., unobserved matrix entries are considered to represent an {\em unknown} relationship.
This is because unobserved diagnoses are not necessarily related to the user's intention, but may result from the user not having received a diagnosis from a medical expert.

To handle unobserved entries in the diagnosis data, we introduce a masking matrix $\mathbf{M}\in \{0,1\}^{M_d\times N}$, where its entry is 1 when the corresponding entry in the diagnosis matrix $\mathbf{P}_d$ is observed and 0 when it is not unobserved.
We modify the objective function in~\autoref{eqn:merged} to incorporate the masking matrix as follows: 

\begin{equation}
\label{eqn:merged_mask}
\begin{split}
\min _{(\mathbf{W}_b, \mathbf{W}_d, \mathbf{W}_s, \mathbf{H})\geq 0} \left\|\mathbf{M} \circ (\mathbf{P}_d-\mathbf{W}_d \mathbf{H})\right\|_F^2 +
\alpha_b&\left\|\mathbf{P}_b-\mathbf{W}_b\mathbf{H}\right\|_F^2 \\
&+\alpha_s\left\|\mathbf{P}_s-\mathbf{W}_s \mathbf{H}\right\|_F^2
.
\end{split}
\end{equation}
As we solve \autoref{eqn:merged}, we use the BCD framework to solve \autoref{eqn:merged_mask}, updating the four factor matrices in each iteration. Updating of $\mathbf{W}_b$ and $\mathbf{W}_s$ can be done in the same way as in \autoref{eqn:optimize_Ws}. However, the updating of $\mathbf{H}$ and $\mathbf{W}_d$ will be different due to the masking matrix.
Considering the effects of the masking matrix, the corresponding subproblems in~\autoref{eqn:optimize_Ws} and~\autoref{eqn:optimize_H} change as follows:
\begin{align*}
\mathbf{W}_d &\leftarrow \argmin_{\mathbf{W}_d \geq 0} \left\|
\mathbf{M} \circ
\left(
\mathbf{P}_d -  \mathbf{W}_d \mathbf{H}
\right)\right\|_F, \\ 
\mathbf{H} &\leftarrow \argmin_{\mathbf{H} \geq 0} \left\|
\begin{bmatrix} 
\mathbf{M} \circ 
\mathbf{P}_d \\
\sqrt{\alpha_b}\mathbf{P}_b \\
\sqrt{\alpha_s}\mathbf{P}_s \end{bmatrix} - \begin{bmatrix} 
\mathbf{M} \circ \left( \mathbf{W}_d \mathbf{H} \right)\\
\sqrt{\alpha_b}\mathbf{W}_b \mathbf{H}\\
\sqrt{\alpha_s}\mathbf{W}_s \mathbf{H}\end{bmatrix}\right\|_F.
\end{align*}

$\mathbf{W}_d$ and $\mathbf{H}$ can be updated row by row and column by column, respectively, using the following update rules:
\begin{align*}
\mathbf{W}_d(j, :) &\leftarrow
\argmin_{\mathbf{W}_d(j, :) \geq 0}\left\|
\mathbf{P}_d(j, :)D(\mathbf{M}(j, :))
-
\mathbf{W}_d(j, :)\mathbf{H} D(\mathbf{M}(j, :))\right\|_F , 
\\
\scalebox{0.95}{$
\mathbf{H}(:, j)$} &\scalebox{0.95}{$\leftarrow$}
\argmin_{\mathbf{H}(:, j) \geq 0}
\scalebox{0.85}{$
\left\|
\begin{bmatrix} 
D(\mathbf{M}(:, j)) 
\mathbf{P}_d(:, j) 
\\
\sqrt{\alpha_b}\mathbf{P}_b(:, j) 
\\
\sqrt{\alpha_s}\mathbf{P}_s(:, j) 
\end{bmatrix} - \begin{bmatrix} 
D(\mathbf{M}(:, j)) \mathbf{W}_d
\\
\sqrt{\alpha_b}\mathbf{W}_b
\\
\sqrt{\alpha_s}\mathbf{W}_s
\end{bmatrix}
\mathbf{H}(:, j)\right\|_F$}, \label{eqn:h_update}
\end{align*}
where $D(\mathbf{z})$ denotes a diagonal matrix constructed from a vector $\mathbf{z}$, where $d_{ii}=z_i$.
\section{Evaluation}
\subsection{Settings}
\subsubsection{Data}
In our evaluation, we used anonymized data collected in 2022 from the Kaiser Permanente Digital (KPD) database and web portal\footnote{\url{https://healthy.kaiserpermanente.org/}}. This dataset includes 30,690 patients' diagnoses, encoded with ICD-10-CM codes, as well as their search and browsing records from the KPD web portal. The data includes 6,521,201 browsing records and 85,245 search entries. All data was anonymized to maintain privacy, in accordance with HIPAA guidelines\footnote{Health Insurance Portability and Accountability Act (\url{https://www.hhs.gov/hipaa/})}.
\vspace{-1mm}
\subsection{Clustering}

\begin{table}[t]
    \centering
    \small
  \setlength{\tabcolsep}{1pt}
    \setlength{\aboverulesep}{1pt}
    \setlength{\belowrulesep}{1pt}
    \caption{Comparison of clustering results. Best results are shown in bold and second best results are underlined.}
\vspace{-1mm}
    \begin{tabular}{L{1.8cm}L{1.0cm}R{2.5cm}R{2.6cm}}
    \toprule
    \textbf{Data Used} & \textbf{Method} & \textbf{Davies-Bouldin \newline index} & \textbf{Silhouette coefficient}\\ \midrule
    Diagnosis & SentBERT & 8.663 & 0.019\\
  & GPT-2 & 2.758 & 0.076\\
    & BioSentVec & 2.458 & 0.121\\
    & NMF & 1.434 & 0.216\\
    
    \midrule
    Diagnosis  & SentBERT & 7.735 & 0.012\\
    + Browsing & GPT-2 & 2.600 & 0.125\\
    & NMF & \underline{1.277} & \underline{0.254}\\
    
    \midrule
    
    Diagnosis & SentBERT & 2.001 & 0.103\\
     + Browsing & GPT-2 & 2.913 & 0.209\\
     + Search & NMF & \textbf{1.144} & \textbf{0.416}\\
   
     \bottomrule
    \end{tabular}
\vspace{-1mm}
    \label{tab:clustering_result}
\end{table}
Our proposed patient profiling method, denoted as nonnegative matrix factorization (NMF), was evaluated using the Davies-Bouldin index \cite{davies1979cluster} and the Silhouette coefficient \cite{rousseeuw1987silhouettes}. Lower values of the former and higher values of the latter indicate superior clustering performance. We compared NMF against standalone usage of text embedding methods: SentenceBERT, GPT-2, and BioSentVec. These embeddings were instrumental in forming our patient profiles. In this evaluation, however, they are also examined for their standalone clustering performance, without being integrated into our NMF method.
Since NMF's output is interpreted as a soft clustering membership~\cite{kuang2015nonnegative}, it allows immediate cluster identification through the index of the maximum value. For the other methods, additional K-means clustering was applied.
All experiments were conducted using 18 clusters, a number determined as optimal using the Gap statistics method \cite{tibshirani2001estimating}.
As presented in Table \ref{tab:clustering_result}, NMF consistently outperformed the other methods across all types of data used.
NMF exhibited the best clustering performance with a Davies-Bouldin index of 1.144 and a Silhouette coefficient of 0.416 when applied to the combined all available data. Interestingly, the performance of NMF consistently improved with each addition of digital data (browsing and search) to the clinical data (diagnosis).
This suggests that combining clinical and digital data can enhance the effectiveness of clustering methods.
\vspace{-1mm}
\begin{table}[t]
\centering
\small  \setlength{\tabcolsep}{1pt}
  \setlength{\tabcolsep}{1pt}
    \setlength{\aboverulesep}{1pt}
    \setlength{\belowrulesep}{1pt}
\caption{Comparative evaluation of recommendation methods. The best performing results are highlighted in bold. (Unit: \%)}
\vspace{-1mm}
\label{tab:auc-scores-xgboost}
\begin{tabular}
{L{2.1cm}R{1.4cm}R{1.3cm}R{1.0cm}R{1.3cm}R{1.1cm}}
\toprule
\textbf{Method} & \textbf{ROC-AUC} & \textbf{Accuracy} & \textbf{Recall} & \textbf{Precision} & \textbf{F1-score} \\
\midrule
HashGNN & 59.38 & 57.14 & 55.90 & 57.30 & 56.59\\
GPT-2 (PCA) & \underline{75.14} & \underline{68.42} & \textbf{71.63} & \underline{67.37} & \underline{69.43} \\
SentBERT (PCA) & 74.29 & 67.56 & 69.65 & 66.82 & 68.21 \\
BioSentVec (PCA) & 63.59 & 59.84 & 67.05 & 58.61 & 62.54 \\
NMF & \textbf{80.92} & \textbf{73.05} & \underline{70.31} & \textbf{74.41} & \textbf{72.30} \\
\bottomrule
\end{tabular}
\vspace{-5mm}
\end{table}
\subsection{Recommendation}
Our method's performance as a representation learning technique for user embeddings was evaluated through a recommendation task related to mental wellness support apps, designed to assist users in managing stress, anxiety, and other mental health issues (see ~\cite{weisel2019standalone}). We transformed the task into a binary classification problem: whether users accessed the download page of a mental wellness support app during the data collection period.
We compared the NMF method against other embedding methods including HashGNN~\cite{tan2020learning}, a graph-based method that does not rely on text information, and other text-based embedding techniques. The models were assessed based on several performance metrics, including ROC-AUC, accuracy, recall, precision, and F1-score.
For comparison purpose, we set the dimensionality of the embedding methods to 128. Both HashGNN and our method NMF were trained with a dimension of 128. The original dimensions of GPT-2, SentenceBERT, and BioSentVec, which were greater than 128, were reduced using Principal Component Analysis (PCA), resulting in embeddings with 128 dimensions for comparison.
We used XGBoost, a gradient boosting classification algorithm ~\cite{chen2016xgboost}.
The experimental results displayed in Table \ref{tab:auc-scores-xgboost} show that our proposed method, NMF, significantly outperforms the baseline methods in terms of ROC-AUC, accuracy, precision, and F1-score. Although the recall of GPT-2 is higher at 71.63\%, the other evaluation metrics are relatively low, supporting the reliability and robustness of our method.
\section{Conclusion}
In this study, we have developed a novel framework that integrates clinical and digital for comprehensive patient profiling. Utilizing constrained low rank approximation techniques, our method simultaneously achieves representation learning and clustering, enhancing performance across tasks. While our focus has been on diagnosis codes, integrating richer data like clinical notes could further enrich the profile. Additionally, exploring its efficacy on relevant public datasets would underscore its wider applicability. Future endeavors can build on these insights to drive further advancements in patient profiling.
\begin{acks}
This research was funded by Kaiser Permanente.
\end{acks}
\bibliographystyle{ACM-Reference-Format}
\bibliography{bibs}


\begin{thebibliography}{41}


\ifx \showCODEN    \undefined \def \showCODEN     #1{\unskip}     \fi
\ifx \showDOI      \undefined \def \showDOI       #1{#1}\fi
\ifx \showISBNx    \undefined \def \showISBNx     #1{\unskip}     \fi
\ifx \showISBNxiii \undefined \def \showISBNxiii  #1{\unskip}     \fi
\ifx \showISSN     \undefined \def \showISSN      #1{\unskip}     \fi
\ifx \showLCCN     \undefined \def \showLCCN      #1{\unskip}     \fi
\ifx \shownote     \undefined \def \shownote      #1{#1}          \fi
\ifx \showarticletitle \undefined \def \showarticletitle #1{#1}   \fi
\ifx \showURL      \undefined \def \showURL       {\relax}        \fi
\providecommand\bibfield[2]{#2}
\providecommand\bibinfo[2]{#2}
\providecommand\natexlab[1]{#1}
\providecommand\showeprint[2][]{arXiv:#2}

\bibitem[Bertsekas(1999)]%
        {bertsekas1997nonlinear}
\bibfield{author}{\bibinfo{person}{D.P. Bertsekas}.}
  \bibinfo{year}{1999}\natexlab{}.
\newblock \bibinfo{booktitle}{\emph{Nonlinear Programming}}.
\newblock \bibinfo{publisher}{Athena Scientific}.
\newblock


\bibitem[Chen et~al\mbox{.}(2019)]%
        {chen2019biosentvec}
\bibfield{author}{\bibinfo{person}{Qingyu Chen}, \bibinfo{person}{Yifan Peng},
  {and} \bibinfo{person}{Zhiyong Lu}.} \bibinfo{year}{2019}\natexlab{}.
\newblock \showarticletitle{BioSentVec: creating sentence embeddings for
  biomedical texts}. In \bibinfo{booktitle}{\emph{2019 IEEE International
  Conference on Healthcare Informatics (ICHI)}}. IEEE, \bibinfo{pages}{1--5}.
\newblock


\bibitem[Chen and Guestrin(2016)]%
        {chen2016xgboost}
\bibfield{author}{\bibinfo{person}{Tianqi Chen} {and} \bibinfo{person}{Carlos
  Guestrin}.} \bibinfo{year}{2016}\natexlab{}.
\newblock \showarticletitle{Xgboost: A scalable tree boosting system}. In
  \bibinfo{booktitle}{\emph{Proceedings of the 22nd acm sigkdd international
  conference on knowledge discovery and data mining}}.
  \bibinfo{pages}{785--794}.
\newblock


\bibitem[Choi et~al\mbox{.}(2016)]%
        {choi2016retain}
\bibfield{author}{\bibinfo{person}{Edward Choi}, \bibinfo{person}{Mohammad~Taha
  Bahadori}, \bibinfo{person}{Jimeng Sun}, \bibinfo{person}{Joshua Kulas},
  \bibinfo{person}{Andy Schuetz}, {and} \bibinfo{person}{Walter Stewart}.}
  \bibinfo{year}{2016}\natexlab{}.
\newblock \showarticletitle{Retain: An interpretable predictive model for
  healthcare using reverse time attention mechanism}.
\newblock \bibinfo{journal}{\emph{Advances in neural information processing
  systems}}  \bibinfo{volume}{29} (\bibinfo{year}{2016}).
\newblock


\bibitem[Chou and Chou(2002)]%
        {chou2002healthcare}
\bibfield{author}{\bibinfo{person}{David~C Chou} {and} \bibinfo{person}{Amy~Y
  Chou}.} \bibinfo{year}{2002}\natexlab{}.
\newblock \showarticletitle{Healthcare information portal: a web technology for
  the healthcare community}.
\newblock \bibinfo{journal}{\emph{Technology in Society}} \bibinfo{volume}{24},
  \bibinfo{number}{3} (\bibinfo{year}{2002}), \bibinfo{pages}{317--330}.
\newblock


\bibitem[Coughlin et~al\mbox{.}(2017)]%
        {coughlin2017patient}
\bibfield{author}{\bibinfo{person}{Steven~S Coughlin},
  \bibinfo{person}{Judith~J Prochaska}, \bibinfo{person}{Lovoria~B Williams},
  \bibinfo{person}{Gina~M Besenyi}, \bibinfo{person}{Vah{\'e} Heboyan},
  \bibinfo{person}{D~Stephen Goggans}, \bibinfo{person}{Wonsuk Yoo}, {and}
  \bibinfo{person}{Gianluca De~Leo}.} \bibinfo{year}{2017}\natexlab{}.
\newblock \showarticletitle{Patient web portals, disease management, and
  primary prevention}.
\newblock \bibinfo{journal}{\emph{Risk management and healthcare policy}}
  (\bibinfo{year}{2017}), \bibinfo{pages}{33--40}.
\newblock


\bibitem[Davies and Bouldin(1979)]%
        {davies1979cluster}
\bibfield{author}{\bibinfo{person}{David~L Davies} {and}
  \bibinfo{person}{Donald~W Bouldin}.} \bibinfo{year}{1979}\natexlab{}.
\newblock \showarticletitle{A cluster separation measure}.
\newblock \bibinfo{journal}{\emph{IEEE transactions on pattern analysis and
  machine intelligence}} \bibinfo{number}{2} (\bibinfo{year}{1979}),
  \bibinfo{pages}{224--227}.
\newblock


\bibitem[Du et~al\mbox{.}(2019)]%
        {du2019hybrid}
\bibfield{author}{\bibinfo{person}{Rundong Du}, \bibinfo{person}{Barry~L.
  Drake}, {and} \bibinfo{person}{Haesun Park}.}
  \bibinfo{year}{2019}\natexlab{}.
\newblock \showarticletitle{Hybrid clustering based on content and connection
  structure using joint nonnegative matrix factorization}.
\newblock \bibinfo{journal}{\emph{J. Glob. Optim.}} \bibinfo{volume}{74},
  \bibinfo{number}{4} (\bibinfo{year}{2019}), \bibinfo{pages}{861--877}.
\newblock
\urldef\tempurl%
\url{https://doi.org/10.1007/s10898-017-0578-x}
\showDOI{\tempurl}


\bibitem[Du et~al\mbox{.}(2017)]%
        {HSymNMF2}
\bibfield{author}{\bibinfo{person}{Rundong Du}, \bibinfo{person}{Da Kuang},
  \bibinfo{person}{Barry Drake}, {and} \bibinfo{person}{Haesun Park}.}
  \bibinfo{year}{2017}\natexlab{}.
\newblock \showarticletitle{Hierarchical Community Detection via Rank-2
  Symmetric Nonnegative Matrix Factorization}.
\newblock \bibinfo{journal}{\emph{Computational Social Networks}}
  \bibinfo{volume}{4} (\bibinfo{date}{12} \bibinfo{year}{2017}),
  \bibinfo{pages}{1 -- 26}.
\newblock
\urldef\tempurl%
\url{https://doi.org/10.1186/s40649-017-0043-5}
\showDOI{\tempurl}


\bibitem[Ekstrand et~al\mbox{.}(2011)]%
        {ekstrand2011collaborative}
\bibfield{author}{\bibinfo{person}{Michael~D Ekstrand}, \bibinfo{person}{John~T
  Riedl}, \bibinfo{person}{Joseph~A Konstan}, {et~al\mbox{.}}}
  \bibinfo{year}{2011}\natexlab{}.
\newblock \showarticletitle{Collaborative filtering recommender systems}.
\newblock \bibinfo{journal}{\emph{Foundations and Trends{\textregistered} in
  Human--Computer Interaction}} \bibinfo{volume}{4}, \bibinfo{number}{2}
  (\bibinfo{year}{2011}), \bibinfo{pages}{81--173}.
\newblock


\bibitem[Henriksson et~al\mbox{.}(2015)]%
        {henriksson2015modeling}
\bibfield{author}{\bibinfo{person}{Aron Henriksson}, \bibinfo{person}{Jing
  Zhao}, \bibinfo{person}{Henrik Bostr{\"o}m}, {and} \bibinfo{person}{Hercules
  Dalianis}.} \bibinfo{year}{2015}\natexlab{}.
\newblock \showarticletitle{Modeling electronic health records in ensembles of
  semantic spaces for adverse drug event detection}. In
  \bibinfo{booktitle}{\emph{2015 IEEE International Conference on
  Bioinformatics and Biomedicine (BIBM)}}. IEEE, \bibinfo{pages}{343--350}.
\newblock


\bibitem[Jain et~al\mbox{.}(2005)]%
        {DBLP:journals/pr/JainNR05}
\bibfield{author}{\bibinfo{person}{Anil~K. Jain}, \bibinfo{person}{Karthik
  Nandakumar}, {and} \bibinfo{person}{Arun Ross}.}
  \bibinfo{year}{2005}\natexlab{}.
\newblock \showarticletitle{Score normalization in multimodal biometric
  systems}.
\newblock \bibinfo{journal}{\emph{Pattern Recognit.}} \bibinfo{volume}{38},
  \bibinfo{number}{12} (\bibinfo{year}{2005}), \bibinfo{pages}{2270--2285}.
\newblock
\urldef\tempurl%
\url{https://doi.org/10.1016/j.patcog.2005.01.012}
\showDOI{\tempurl}


\bibitem[Kim et~al\mbox{.}(2015)]%
        {kim2015simultaneous}
\bibfield{author}{\bibinfo{person}{Hannah Kim}, \bibinfo{person}{Jaegul Choo},
  \bibinfo{person}{Jingu Kim}, \bibinfo{person}{Chandan~K. Reddy}, {and}
  \bibinfo{person}{Haesun Park}.} \bibinfo{year}{2015}\natexlab{}.
\newblock \showarticletitle{Simultaneous Discovery of Common and Discriminative
  Topics via Joint Nonnegative Matrix Factorization}. In
  \bibinfo{booktitle}{\emph{Proceedings of the 21th {ACM} {SIGKDD}
  International Conference on Knowledge Discovery and Data Mining, Sydney, NSW,
  Australia, August 10-13, 2015}}, \bibfield{editor}{\bibinfo{person}{Longbing
  Cao}, \bibinfo{person}{Chengqi Zhang}, \bibinfo{person}{Thorsten Joachims},
  \bibinfo{person}{Geoffrey~I. Webb}, \bibinfo{person}{Dragos~D. Margineantu},
  {and} \bibinfo{person}{Graham Williams}} (Eds.). \bibinfo{publisher}{{ACM}},
  \bibinfo{pages}{567--576}.
\newblock
\urldef\tempurl%
\url{https://doi.org/10.1145/2783258.2783338}
\showDOI{\tempurl}


\bibitem[Kim and Park(2008a)]%
        {kim2008nonnegative}
\bibfield{author}{\bibinfo{person}{Hyunsoo Kim} {and} \bibinfo{person}{Haesun
  Park}.} \bibinfo{year}{2008}\natexlab{a}.
\newblock \showarticletitle{Nonnegative Matrix Factorization Based on
  Alternating Nonnegativity Constrained Least Squares and Active Set Method}.
\newblock \bibinfo{journal}{\emph{{SIAM} J. Matrix Anal. Appl.}}
  \bibinfo{volume}{30}, \bibinfo{number}{2} (\bibinfo{year}{2008}),
  \bibinfo{pages}{713--730}.
\newblock


\bibitem[Kim et~al\mbox{.}(2014)]%
        {kim2014algorithms}
\bibfield{author}{\bibinfo{person}{Jingu Kim}, \bibinfo{person}{Yunlong He},
  {and} \bibinfo{person}{Haesun Park}.} \bibinfo{year}{2014}\natexlab{}.
\newblock \showarticletitle{Algorithms for nonnegative matrix and tensor
  factorizations: a unified view based on block coordinate descent framework}.
\newblock \bibinfo{journal}{\emph{J. Glob. Optim.}} \bibinfo{volume}{58},
  \bibinfo{number}{2} (\bibinfo{year}{2014}), \bibinfo{pages}{285--319}.
\newblock


\bibitem[Kim and Park(2008b)]%
        {kim2008sparse}
\bibfield{author}{\bibinfo{person}{Jingu Kim} {and} \bibinfo{person}{Haesun
  Park}.} \bibinfo{year}{2008}\natexlab{b}.
\newblock \bibinfo{booktitle}{\emph{Sparse nonnegative matrix factorization for
  clustering}}.
\newblock \bibinfo{type}{{T}echnical {R}eport}. \bibinfo{institution}{Georgia
  Institute of Technology}.
\newblock


\bibitem[Kim and Park(2011)]%
        {kim2011fast}
\bibfield{author}{\bibinfo{person}{Jingu Kim} {and} \bibinfo{person}{Haesun
  Park}.} \bibinfo{year}{2011}\natexlab{}.
\newblock \showarticletitle{Fast Nonnegative Matrix Factorization: An
  Active-Set-Like Method and Comparisons}.
\newblock \bibinfo{journal}{\emph{{SIAM} J. Sci. Comput.}}
  \bibinfo{volume}{33}, \bibinfo{number}{6} (\bibinfo{year}{2011}),
  \bibinfo{pages}{3261--3281}.
\newblock
\urldef\tempurl%
\url{https://doi.org/10.1137/110821172}
\showDOI{\tempurl}


\bibitem[Kuang et~al\mbox{.}(2015a)]%
        {kuang2015nonnegative}
\bibfield{author}{\bibinfo{person}{Da Kuang}, \bibinfo{person}{Jaegul Choo},
  {and} \bibinfo{person}{Haesun Park}.} \bibinfo{year}{2015}\natexlab{a}.
\newblock \showarticletitle{Nonnegative matrix factorization for interactive
  topic modeling and document clustering}.
\newblock \bibinfo{journal}{\emph{Partitional clustering algorithms}}
  (\bibinfo{year}{2015}), \bibinfo{pages}{215--243}.
\newblock


\bibitem[Kuang et~al\mbox{.}(2015b)]%
        {SymNMF_Jogo}
\bibfield{author}{\bibinfo{person}{Da Kuang}, \bibinfo{person}{Sangwoon Yun},
  {and} \bibinfo{person}{Haesun Park}.} \bibinfo{year}{2015}\natexlab{b}.
\newblock \showarticletitle{SymNMF: Nonnegative low-rank approximation of a
  similarity matrix for graph clustering}.
\newblock \bibinfo{journal}{\emph{Journal of Global Optimization}}
  \bibinfo{volume}{62} (\bibinfo{date}{07} \bibinfo{year}{2015}).
\newblock
\urldef\tempurl%
\url{https://doi.org/10.1007/s10898-014-0247-2}
\showDOI{\tempurl}


\bibitem[Le and Mikolov(2014)]%
        {le2014distributed}
\bibfield{author}{\bibinfo{person}{Quoc~V. Le} {and}
  \bibinfo{person}{Tom{\'{a}}s Mikolov}.} \bibinfo{year}{2014}\natexlab{}.
\newblock \showarticletitle{Distributed Representations of Sentences and
  Documents}. In \bibinfo{booktitle}{\emph{Proceedings of the 31th
  International Conference on Machine Learning, {ICML} 2014, Beijing, China,
  21-26 June 2014}} \emph{(\bibinfo{series}{{JMLR} Workshop and Conference
  Proceedings}, Vol.~\bibinfo{volume}{32})}. \bibinfo{publisher}{JMLR.org},
  \bibinfo{pages}{1188--1196}.
\newblock
\urldef\tempurl%
\url{http://proceedings.mlr.press/v32/le14.html}
\showURL{%
\tempurl}


\bibitem[Mikolov et~al\mbox{.}(2013a)]%
        {DBLP:conf/nips/MikolovSCCD13}
\bibfield{author}{\bibinfo{person}{Tom{\'{a}}s Mikolov}, \bibinfo{person}{Ilya
  Sutskever}, \bibinfo{person}{Kai Chen}, \bibinfo{person}{Gregory~S. Corrado},
  {and} \bibinfo{person}{Jeffrey Dean}.} \bibinfo{year}{2013}\natexlab{a}.
\newblock \showarticletitle{Distributed Representations of Words and Phrases
  and their Compositionality}. In \bibinfo{booktitle}{\emph{Advances in Neural
  Information Processing Systems 26: 27th Annual Conference on Neural
  Information Processing Systems 2013. Proceedings of a meeting held December
  5-8, 2013, Lake Tahoe, Nevada, United States}},
  \bibfield{editor}{\bibinfo{person}{Christopher J.~C. Burges},
  \bibinfo{person}{L{\'{e}}on Bottou}, \bibinfo{person}{Zoubin Ghahramani},
  {and} \bibinfo{person}{Kilian~Q. Weinberger}} (Eds.).
  \bibinfo{pages}{3111--3119}.
\newblock


\bibitem[Mikolov et~al\mbox{.}(2013b)]%
        {mikolov2013distributed}
\bibfield{author}{\bibinfo{person}{Tom{\'{a}}s Mikolov}, \bibinfo{person}{Ilya
  Sutskever}, \bibinfo{person}{Kai Chen}, \bibinfo{person}{Gregory~S. Corrado},
  {and} \bibinfo{person}{Jeffrey Dean}.} \bibinfo{year}{2013}\natexlab{b}.
\newblock \showarticletitle{Distributed Representations of Words and Phrases
  and their Compositionality}. In \bibinfo{booktitle}{\emph{Advances in Neural
  Information Processing Systems 26: 27th Annual Conference on Neural
  Information Processing Systems 2013. Proceedings of a meeting held December
  5-8, 2013, Lake Tahoe, Nevada, United States}},
  \bibfield{editor}{\bibinfo{person}{Christopher J.~C. Burges},
  \bibinfo{person}{L{\'{e}}on Bottou}, \bibinfo{person}{Zoubin Ghahramani},
  {and} \bibinfo{person}{Kilian~Q. Weinberger}} (Eds.).
  \bibinfo{pages}{3111--3119}.
\newblock


\bibitem[Moody(2005)]%
        {moody2005health}
\bibfield{author}{\bibinfo{person}{Linda~E Moody}.}
  \bibinfo{year}{2005}\natexlab{}.
\newblock \showarticletitle{E-health web portals: delivering holistic
  healthcare and making home the point of care}.
\newblock \bibinfo{journal}{\emph{Holistic nursing practice}}
  \bibinfo{volume}{19}, \bibinfo{number}{4} (\bibinfo{year}{2005}),
  \bibinfo{pages}{156--160}.
\newblock


\bibitem[Nickel et~al\mbox{.}(2016)]%
        {DBLP:journals/pieee/Nickel0TG16}
\bibfield{author}{\bibinfo{person}{Maximilian Nickel}, \bibinfo{person}{Kevin
  Murphy}, \bibinfo{person}{Volker Tresp}, {and} \bibinfo{person}{Evgeniy
  Gabrilovich}.} \bibinfo{year}{2016}\natexlab{}.
\newblock \showarticletitle{A Review of Relational Machine Learning for
  Knowledge Graphs}.
\newblock \bibinfo{journal}{\emph{Proc. {IEEE}}} \bibinfo{volume}{104},
  \bibinfo{number}{1} (\bibinfo{year}{2016}), \bibinfo{pages}{11--33}.
\newblock
\urldef\tempurl%
\url{https://doi.org/10.1109/JPROC.2015.2483592}
\showDOI{\tempurl}


\bibitem[Organization et~al\mbox{.}(1992)]%
        {world1992icd}
\bibfield{author}{\bibinfo{person}{World~Health Organization} {et~al\mbox{.}}}
  \bibinfo{year}{1992}\natexlab{}.
\newblock \bibinfo{booktitle}{\emph{The ICD-10 classification of mental and
  behavioural disorders: clinical descriptions and diagnostic guidelines}}.
\newblock \bibinfo{publisher}{World Health Organization}.
\newblock


\bibitem[Pretschner and Gauch(1999)]%
        {DBLP:conf/ictai/PretschnerG99}
\bibfield{author}{\bibinfo{person}{Alexander Pretschner} {and}
  \bibinfo{person}{Susan Gauch}.} \bibinfo{year}{1999}\natexlab{}.
\newblock \showarticletitle{Ontology Based Personalized Search}. In
  \bibinfo{booktitle}{\emph{11th {IEEE} International Conference on Tools with
  Artificial Intelligence, {ICTAI} '99, Chicago, Illinois, USA, November 8-10,
  1999}}. \bibinfo{publisher}{{IEEE} Computer Society},
  \bibinfo{pages}{391--398}.
\newblock
\urldef\tempurl%
\url{https://doi.org/10.1109/TAI.1999.809829}
\showDOI{\tempurl}


\bibitem[Radford et~al\mbox{.}(2019)]%
        {radford2019language}
\bibfield{author}{\bibinfo{person}{Alec Radford}, \bibinfo{person}{Jeffrey Wu},
  \bibinfo{person}{Rewon Child}, \bibinfo{person}{David Luan},
  \bibinfo{person}{Dario Amodei}, \bibinfo{person}{Ilya Sutskever},
  {et~al\mbox{.}}} \bibinfo{year}{2019}\natexlab{}.
\newblock \showarticletitle{Language models are unsupervised multitask
  learners}.
\newblock \bibinfo{journal}{\emph{OpenAI blog}} \bibinfo{volume}{1},
  \bibinfo{number}{8} (\bibinfo{year}{2019}), \bibinfo{pages}{9}.
\newblock


\bibitem[Reimers and Gurevych(2019)]%
        {reimers2019sentence}
\bibfield{author}{\bibinfo{person}{Nils Reimers} {and} \bibinfo{person}{Iryna
  Gurevych}.} \bibinfo{year}{2019}\natexlab{}.
\newblock \showarticletitle{Sentence-BERT: Sentence Embeddings using Siamese
  BERT-Networks}. In \bibinfo{booktitle}{\emph{Proceedings of the 2019
  Conference on Empirical Methods in Natural Language Processing and the 9th
  International Joint Conference on Natural Language Processing, {EMNLP-IJCNLP}
  2019, Hong Kong, China, November 3-7, 2019}},
  \bibfield{editor}{\bibinfo{person}{Kentaro Inui}, \bibinfo{person}{Jing
  Jiang}, \bibinfo{person}{Vincent Ng}, {and} \bibinfo{person}{Xiaojun Wan}}
  (Eds.). \bibinfo{publisher}{Association for Computational Linguistics},
  \bibinfo{pages}{3980--3990}.
\newblock
\urldef\tempurl%
\url{https://doi.org/10.18653/v1/D19-1410}
\showDOI{\tempurl}


\bibitem[Reiter(1977)]%
        {DBLP:conf/adbt/Reiter77}
\bibfield{author}{\bibinfo{person}{Raymond Reiter}.}
  \bibinfo{year}{1977}\natexlab{}.
\newblock \showarticletitle{On Closed World Data Bases}. In
  \bibinfo{booktitle}{\emph{Logic and Data Bases, Symposium on Logic and Data
  Bases, Centre d'{\'{e}}tudes et de recherches de Toulouse, France, 1977}}
  \emph{(\bibinfo{series}{Advances in Data Base Theory})},
  \bibfield{editor}{\bibinfo{person}{Herv{\'{e}} Gallaire} {and}
  \bibinfo{person}{Jack Minker}} (Eds.). \bibinfo{publisher}{Plemum Press},
  \bibinfo{address}{New York}, \bibinfo{pages}{55--76}.
\newblock
\urldef\tempurl%
\url{https://doi.org/10.1007/978-1-4684-3384-5\_3}
\showDOI{\tempurl}


\bibitem[Rousseeuw(1987)]%
        {rousseeuw1987silhouettes}
\bibfield{author}{\bibinfo{person}{Peter~J Rousseeuw}.}
  \bibinfo{year}{1987}\natexlab{}.
\newblock \showarticletitle{Silhouettes: a graphical aid to the interpretation
  and validation of cluster analysis}.
\newblock \bibinfo{journal}{\emph{Journal of computational and applied
  mathematics}}  \bibinfo{volume}{20} (\bibinfo{year}{1987}),
  \bibinfo{pages}{53--65}.
\newblock


\bibitem[Sammut and Webb(2010)]%
        {DBLP:reference/ml/2010}
\bibfield{editor}{\bibinfo{person}{Claude Sammut} {and}
  \bibinfo{person}{Geoffrey~I. Webb}} (Eds.). \bibinfo{year}{2010}\natexlab{}.
\newblock \bibinfo{booktitle}{\emph{Encyclopedia of Machine Learning}}.
\newblock \bibinfo{publisher}{Springer}.
\newblock
\showISBNx{978-0-387-30768-8}
\urldef\tempurl%
\url{https://doi.org/10.1007/978-0-387-30164-8}
\showDOI{\tempurl}


\bibitem[Shen et~al\mbox{.}(2005)]%
        {DBLP:conf/cikm/ShenTZ05}
\bibfield{author}{\bibinfo{person}{Xuehua Shen}, \bibinfo{person}{Bin Tan},
  {and} \bibinfo{person}{ChengXiang Zhai}.} \bibinfo{year}{2005}\natexlab{}.
\newblock \showarticletitle{Implicit user modeling for personalized search}. In
  \bibinfo{booktitle}{\emph{Proceedings of the 2005 {ACM} {CIKM} International
  Conference on Information and Knowledge Management, Bremen, Germany, October
  31 - November 5, 2005}}, \bibfield{editor}{\bibinfo{person}{Otthein Herzog},
  \bibinfo{person}{Hans{-}J{\"{o}}rg Schek}, \bibinfo{person}{Norbert Fuhr},
  \bibinfo{person}{Abdur Chowdhury}, {and} \bibinfo{person}{Wilfried Teiken}}
  (Eds.). \bibinfo{publisher}{{ACM}}, \bibinfo{pages}{824--831}.
\newblock
\urldef\tempurl%
\url{https://doi.org/10.1145/1099554.1099747}
\showDOI{\tempurl}


\bibitem[Stratigi et~al\mbox{.}(2020)]%
        {stratigi2020multidimensional}
\bibfield{author}{\bibinfo{person}{Maria Stratigi}, \bibinfo{person}{Haridimos
  Kondylakis}, {and} \bibinfo{person}{Kostas Stefanidis}.}
  \bibinfo{year}{2020}\natexlab{}.
\newblock \showarticletitle{Multidimensional group recommendations in the
  health domain}.
\newblock \bibinfo{journal}{\emph{Algorithms}} \bibinfo{volume}{13},
  \bibinfo{number}{3} (\bibinfo{year}{2020}), \bibinfo{pages}{54}.
\newblock


\bibitem[Sugiyama et~al\mbox{.}(2004)]%
        {DBLP:conf/www/SugiyamaHY04}
\bibfield{author}{\bibinfo{person}{Kazunari Sugiyama}, \bibinfo{person}{Kenji
  Hatano}, {and} \bibinfo{person}{Masatoshi Yoshikawa}.}
  \bibinfo{year}{2004}\natexlab{}.
\newblock \showarticletitle{Adaptive web search based on user profile
  constructed without any effort from users}. In
  \bibinfo{booktitle}{\emph{Proceedings of the 13th international conference on
  World Wide Web, {WWW} 2004, New York, NY, USA, May 17-20, 2004}},
  \bibfield{editor}{\bibinfo{person}{Stuart~I. Feldman}, \bibinfo{person}{Mike
  Uretsky}, \bibinfo{person}{Marc Najork}, {and} \bibinfo{person}{Craig~E.
  Wills}} (Eds.). \bibinfo{publisher}{{ACM}}, \bibinfo{pages}{675--684}.
\newblock
\urldef\tempurl%
\url{https://doi.org/10.1145/988672.988764}
\showDOI{\tempurl}


\bibitem[Tan et~al\mbox{.}(2006)]%
        {DBLP:conf/kdd/TanSZ06}
\bibfield{author}{\bibinfo{person}{Bin Tan}, \bibinfo{person}{Xuehua Shen},
  {and} \bibinfo{person}{ChengXiang Zhai}.} \bibinfo{year}{2006}\natexlab{}.
\newblock \showarticletitle{Mining long-term search history to improve search
  accuracy}. In \bibinfo{booktitle}{\emph{Proceedings of the Twelfth {ACM}
  {SIGKDD} International Conference on Knowledge Discovery and Data Mining,
  Philadelphia, PA, USA, August 20-23, 2006}},
  \bibfield{editor}{\bibinfo{person}{Tina Eliassi{-}Rad},
  \bibinfo{person}{Lyle~H. Ungar}, \bibinfo{person}{Mark Craven}, {and}
  \bibinfo{person}{Dimitrios Gunopulos}} (Eds.). \bibinfo{publisher}{{ACM}},
  \bibinfo{pages}{718--723}.
\newblock
\urldef\tempurl%
\url{https://doi.org/10.1145/1150402.1150493}
\showDOI{\tempurl}


\bibitem[Tan et~al\mbox{.}(2020)]%
        {tan2020learning}
\bibfield{author}{\bibinfo{person}{Qiaoyu Tan}, \bibinfo{person}{Ninghao Liu},
  \bibinfo{person}{Xing Zhao}, \bibinfo{person}{Hongxia Yang},
  \bibinfo{person}{Jingren Zhou}, {and} \bibinfo{person}{Xia Hu}.}
  \bibinfo{year}{2020}\natexlab{}.
\newblock \showarticletitle{Learning to hash with graph neural networks for
  recommender systems}. In \bibinfo{booktitle}{\emph{Proceedings of The Web
  Conference 2020}}. \bibinfo{pages}{1988--1998}.
\newblock


\bibitem[Tan et~al\mbox{.}(2022)]%
        {tan2022metacare++}
\bibfield{author}{\bibinfo{person}{Yanchao Tan}, \bibinfo{person}{Carl Yang},
  \bibinfo{person}{Xiangyu Wei}, \bibinfo{person}{Chaochao Chen},
  \bibinfo{person}{Weiming Liu}, \bibinfo{person}{Longfei Li},
  \bibinfo{person}{Jun Zhou}, {and} \bibinfo{person}{Xiaolin Zheng}.}
  \bibinfo{year}{2022}\natexlab{}.
\newblock \showarticletitle{Metacare++: Meta-learning with hierarchical
  subtyping for cold-start diagnosis prediction in healthcare data}. In
  \bibinfo{booktitle}{\emph{Proceedings of the 45th International ACM SIGIR
  Conference on Research and Development in Information Retrieval}}.
  \bibinfo{pages}{449--459}.
\newblock


\bibitem[Tanbeer and Sykes(2021)]%
        {tanbeer2021myhealthportal}
\bibfield{author}{\bibinfo{person}{Syed~K Tanbeer} {and}
  \bibinfo{person}{Edward~R Sykes}.} \bibinfo{year}{2021}\natexlab{}.
\newblock \showarticletitle{MyHealthPortal--A web-based e-Healthcare web portal
  for out-of-hospital patient care}.
\newblock \bibinfo{journal}{\emph{Digital Health}}  \bibinfo{volume}{7}
  (\bibinfo{year}{2021}), \bibinfo{pages}{2055207621989194}.
\newblock


\bibitem[Tibshirani et~al\mbox{.}(2001)]%
        {tibshirani2001estimating}
\bibfield{author}{\bibinfo{person}{Robert Tibshirani},
  \bibinfo{person}{Guenther Walther}, {and} \bibinfo{person}{Trevor Hastie}.}
  \bibinfo{year}{2001}\natexlab{}.
\newblock \showarticletitle{Estimating the number of clusters in a data set via
  the gap statistic}.
\newblock \bibinfo{journal}{\emph{Journal of the Royal Statistical Society:
  Series B (Statistical Methodology)}} \bibinfo{volume}{63},
  \bibinfo{number}{2} (\bibinfo{year}{2001}), \bibinfo{pages}{411--423}.
\newblock


\bibitem[Weisel et~al\mbox{.}(2019)]%
        {weisel2019standalone}
\bibfield{author}{\bibinfo{person}{Kiona~K Weisel}, \bibinfo{person}{Lukas~M
  Fuhrmann}, \bibinfo{person}{Matthias Berking}, \bibinfo{person}{Harald
  Baumeister}, \bibinfo{person}{Pim Cuijpers}, {and} \bibinfo{person}{David~D
  Ebert}.} \bibinfo{year}{2019}\natexlab{}.
\newblock \showarticletitle{Standalone smartphone apps for mental health—a
  systematic review and meta-analysis}.
\newblock \bibinfo{journal}{\emph{NPJ digital medicine}} \bibinfo{volume}{2},
  \bibinfo{number}{1} (\bibinfo{year}{2019}), \bibinfo{pages}{118}.
\newblock


\bibitem[Whang et~al\mbox{.}(2020)]%
        {whang2020mega}
\bibfield{author}{\bibinfo{person}{Joyce~Jiyoung Whang},
  \bibinfo{person}{Rundong Du}, \bibinfo{person}{Sangwon Jung},
  \bibinfo{person}{Geon Lee}, \bibinfo{person}{Barry Drake},
  \bibinfo{person}{Qingqing Liu}, \bibinfo{person}{Seonggoo Kang}, {and}
  \bibinfo{person}{Haesun Park}.} \bibinfo{year}{2020}\natexlab{}.
\newblock \showarticletitle{MEGA: Multi-view semi-supervised clustering of
  hypergraphs}.
\newblock \bibinfo{journal}{\emph{Proceedings of the VLDB Endowment}}
  \bibinfo{volume}{13}, \bibinfo{number}{5} (\bibinfo{year}{2020}),
  \bibinfo{pages}{698--711}.
\newblock


\end{thebibliography}

\end{document}